\documentclass[a4paper,journal,12pt,onecolumn,final]{IEEEtran}%
\usepackage{amsfonts}
\usepackage{amsmath}
\usepackage{amssymb}
\usepackage{graphicx}%
\providecommand{\U}[1]{\protect \rule{.1in}{.1in}}

\begin{document}

\title{A Preliminary Study on Optimal Placement of Cameras}
\author{Lin Xu \thanks{Lin Xu is with Beijing 101 Middle School, Beijing 100091,
People's Republic of China, Email: linxu\_2002@163.com}}
\maketitle

\begin{abstract}
This paper primarily focuses on figuring out the best array of cameras, or
visual sensors, so that such a placement enables the maximum utilization of
these visual sensors. Maximizing the utilization of these cameras can convert
to another problem that is simpler for the formulation, that is maximizing the
total coverage with these cameras. To solve the problem, the coverage problem
is first defined subject to the capabilities and limits of cameras. Then, each
camera' pose is analyzed for the best arrangement.

\end{abstract}


\section{Introduction}

Cameras are ubiquitous in our lives, as we can almost see it everywhere we go.
Streets, restaurants, offices, schools, areas where people go and stay, are
under these "eyes." Cameras, or in a broad way of definition, visual sensors,
are widely utilized for video surveillance, recording, military uses, and so
on. Given the area that the cameras need to watch, if all cameras are the
same, the size of the area is proportional to the number of cameras.
Therefore, the cost of setting cameras is positively related to the number of
cameras employed. Sometimes, designers make coverage mistakes, ineffective
setting visual sensors, as shown below. In this case, optimal placement of
cameras is critical. \begin{figure}[h]
\begin{center}
\includegraphics[scale=0.8 ]{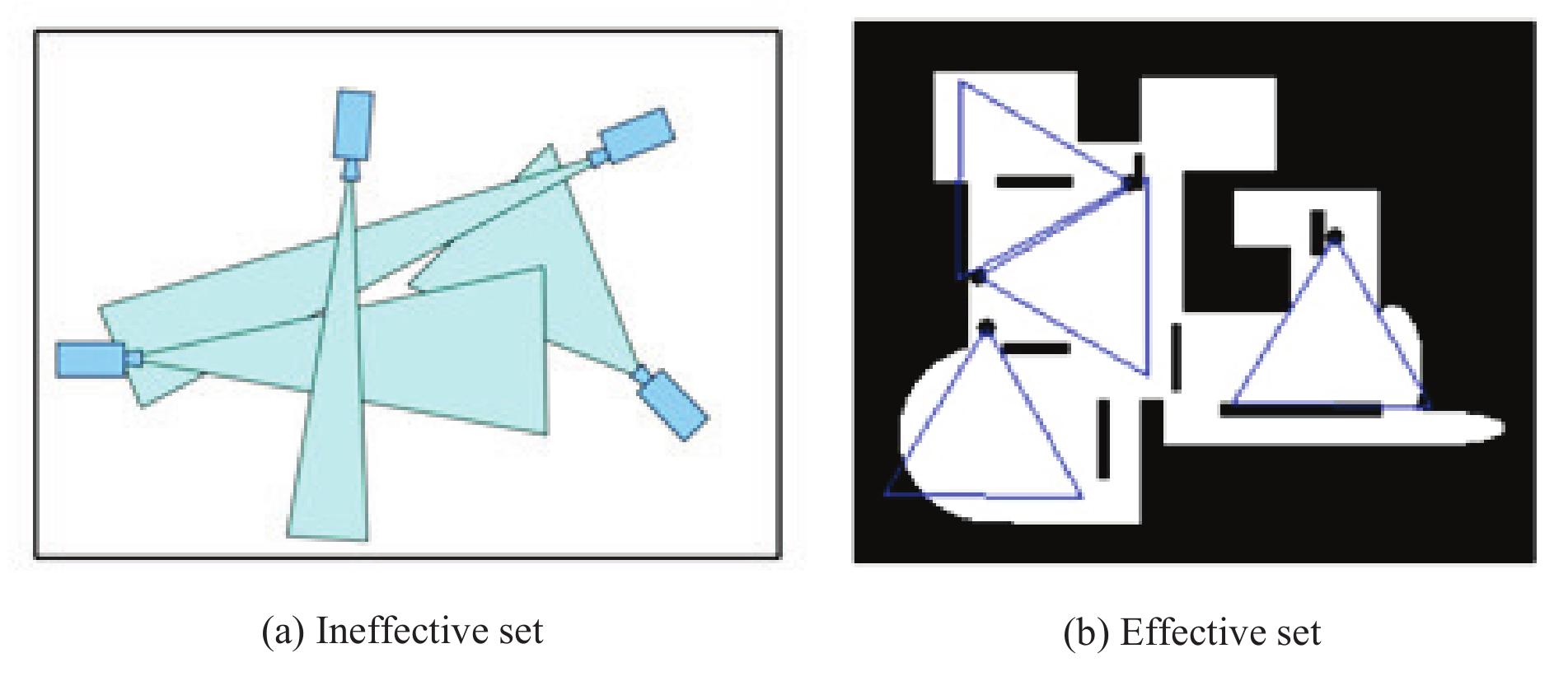}
\end{center}
\caption{Different camera set}%
\end{figure}

In 1973, mathematician Victor Klee proposed the art gallery problem, a
real-world problem trying to find a minimum number of guards who together can
observe the whole gallery. These guards are at fixed positions, and the
gallery shapes like a polygon. Similarly, there is also the Floodlight
illumination problem, coping with the illumination of planar regions by light
sources. For more information, please refer to [1],[2]. However, these
solutions contain unrealistic factors, such as unlimited field of view, the
infinite depth of field, making these algorithms unsuitable for most
real-world computer vision applications.

\section{Problem Formulation}

In this paper, as shown in Figure 2, the problem of optimal camera placement
for a given region is formulated. We focus on the static camera placement
problem, where the goal is to determine optimal position and number of cameras
for a region to be observed, given a set of task-specific constraints, and a
set of possible cameras to use in the layout. \begin{figure}[h]
\begin{center}
\includegraphics[scale=0.8 ]{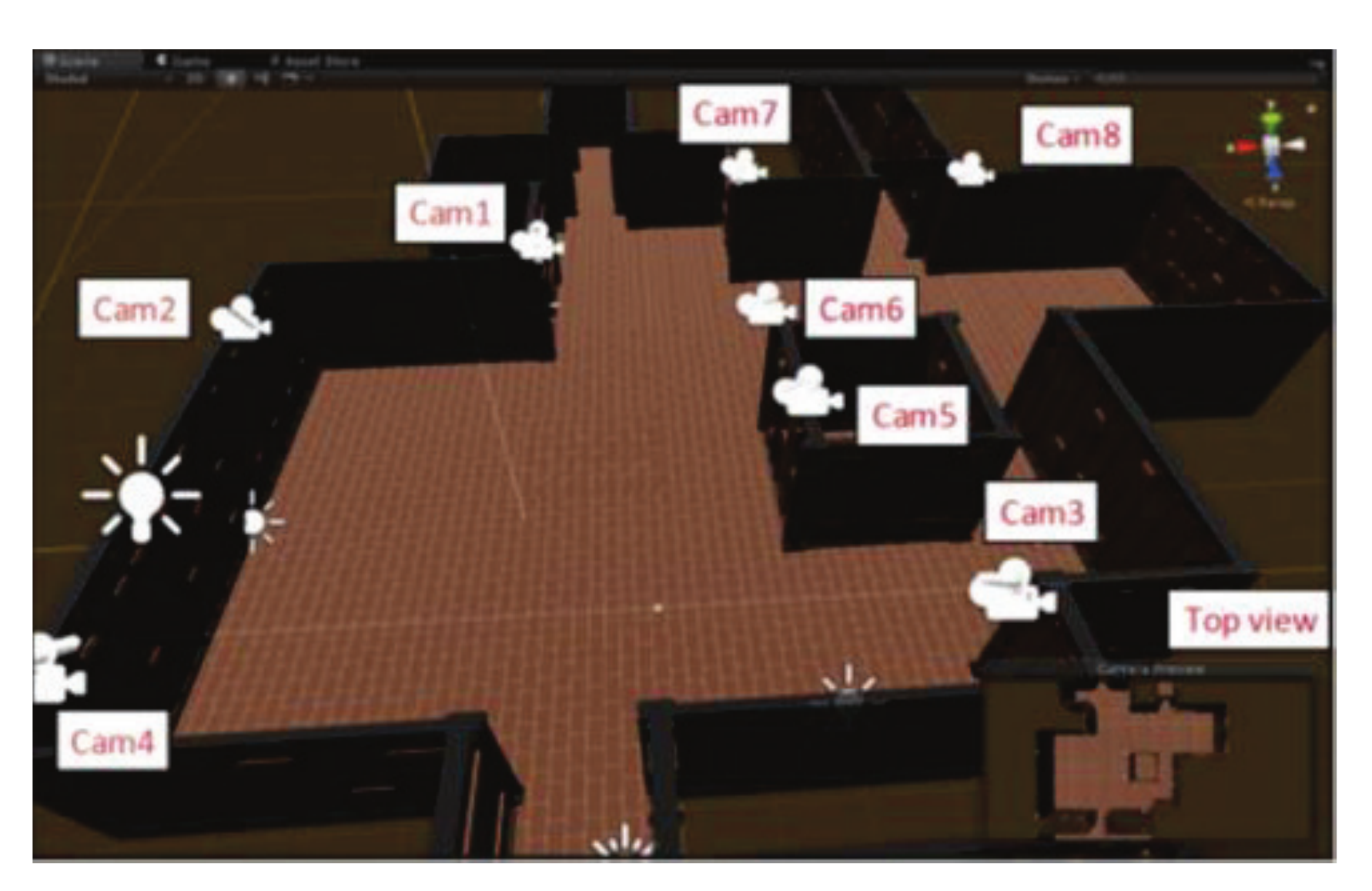}
\end{center}
\caption{Camera placement problem}%
\end{figure}

For simplicity, we first assume that the detection range of each visual sensor
is unbounded. However, there are several constraints needed to consider: (1)
the number of cameras that given, (2) the each camera's limited field-of-view,
(3) fixed position, (4) the room without obstacles. The optimal placement is
defined as the cameras are placed so that they cover as much space as possible.

The field-of-view of a camera can be described by a triangle, as shown in
Figure 3. There are three parameters, with $d$ representing the length of
sight, $\varphi$ determining the pose of a camera, and $2\alpha$ defining the
field-of-view angle.\begin{figure}[h]
\begin{center}
\includegraphics[scale=0.8 ]{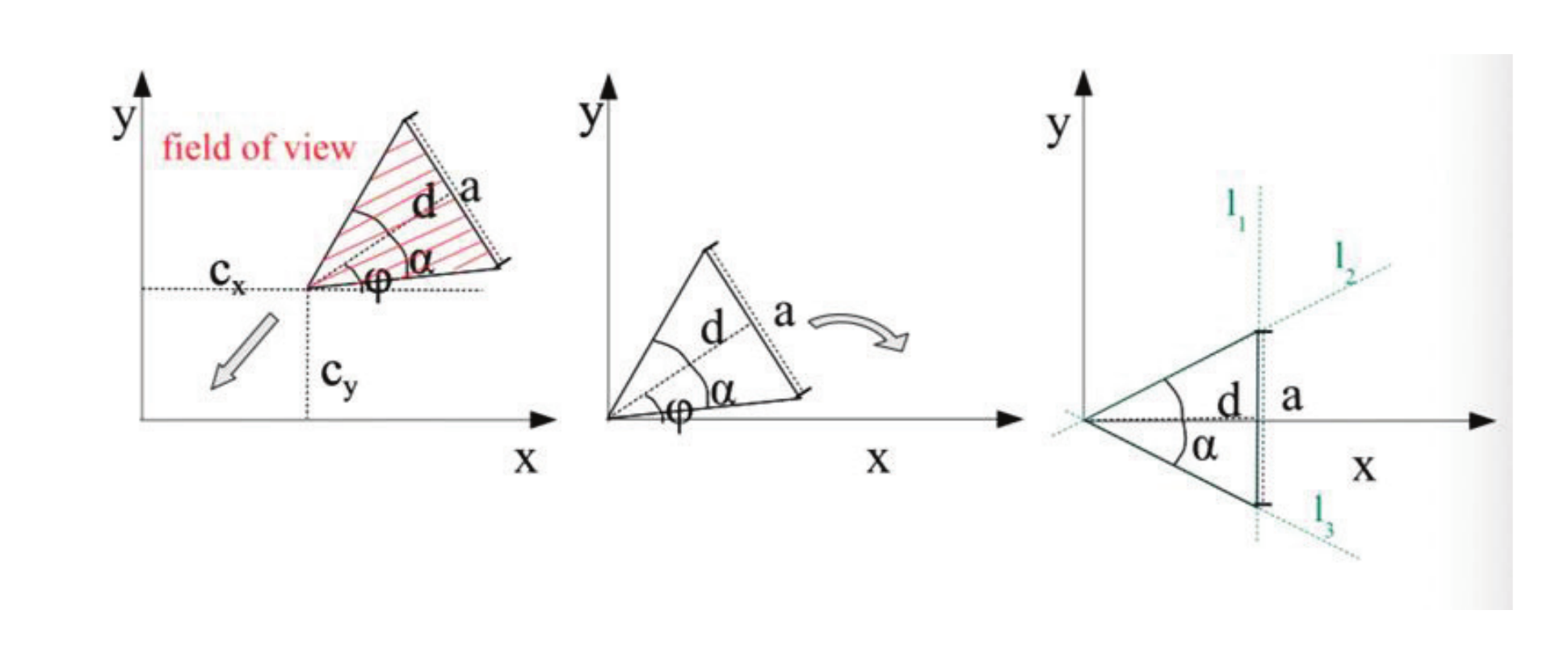}
\end{center}
\caption{Different camera set}%
\end{figure}

\section{Main Result}

For a cuboid 3D space, consider the camera working in 2D space such as 2D
space from the side view and 2D space from the top view as shown in Figure 4.
As for the coverage problem, we can claim that if a place is covered by both
the side view and the top view of a camera, then it can be covered by the
field-of-view angle of the camera in 3D space. With this result, we divide the
coverage problem in 3D space into two subproblems, namely from the top view
and side view. \begin{figure}[h]
\begin{center}
\includegraphics[scale=0.8 ]{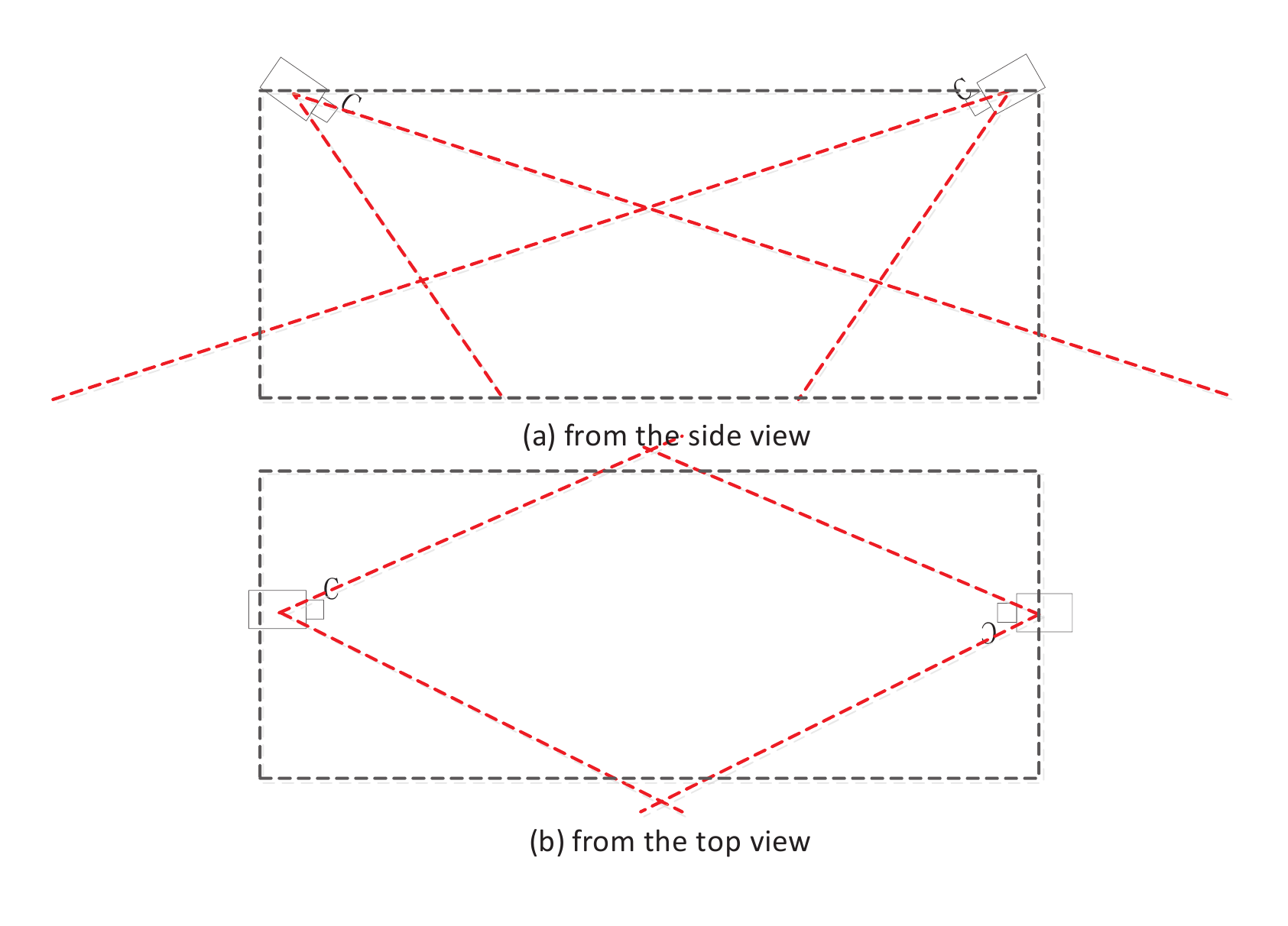}
\end{center}
\caption{2D space coverage scenario}%
\end{figure}

\subsection{From the top view}

From the top view, the best arrangement should be that all spaces are covered
ideally, and the overlapped areas should be minimal. Due to such an analysis,
the cameras should be placed in a staggered way such that these visual sensors
would not monitor a space multiple times, shown in Figure 5. \begin{figure}[h]
\begin{center}
\includegraphics[scale=0.5 ]{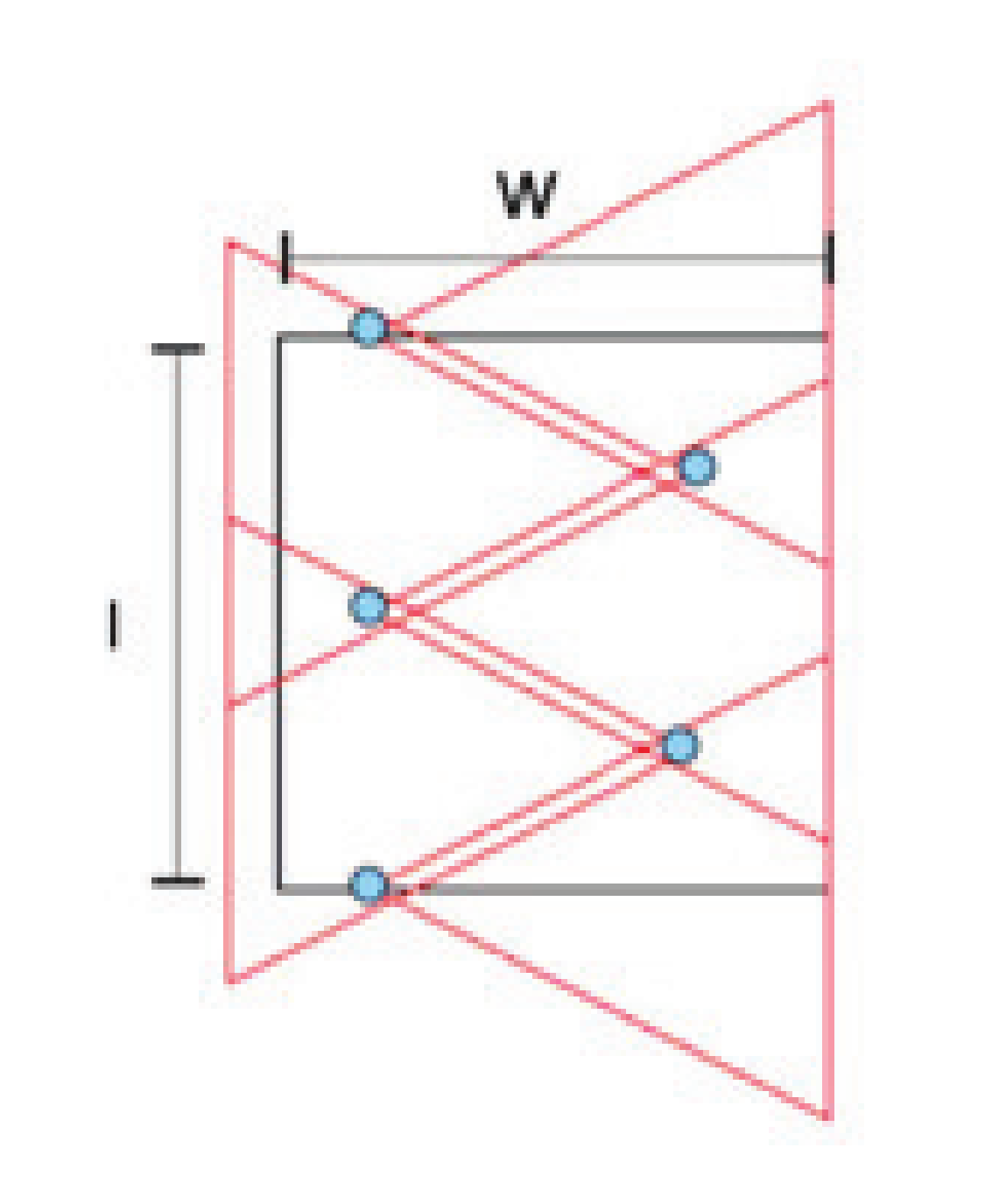}
\end{center}
\caption{Staggered arrangement of cameras}%
\end{figure}

\textbf{Proposition 1}. \textit{As shown in Figure 6, given that cameras with
the same field-of-view angle 2 and unbounded the depth of view, the staggered arrangement is optimal for a rectangular space.}

\textbf{Proof}. In order to get the maximum coverage, the pose of a camera
should be changed such that all field-of-view angle is included in the given
space, as shown in Figure 6. Six cameras are labelled by A to F, and their
field-of-view angles also correspond to1 to 6 in Figure 6,
respectively.\begin{figure}[h]
\begin{center}
\includegraphics[scale=0.8 ]{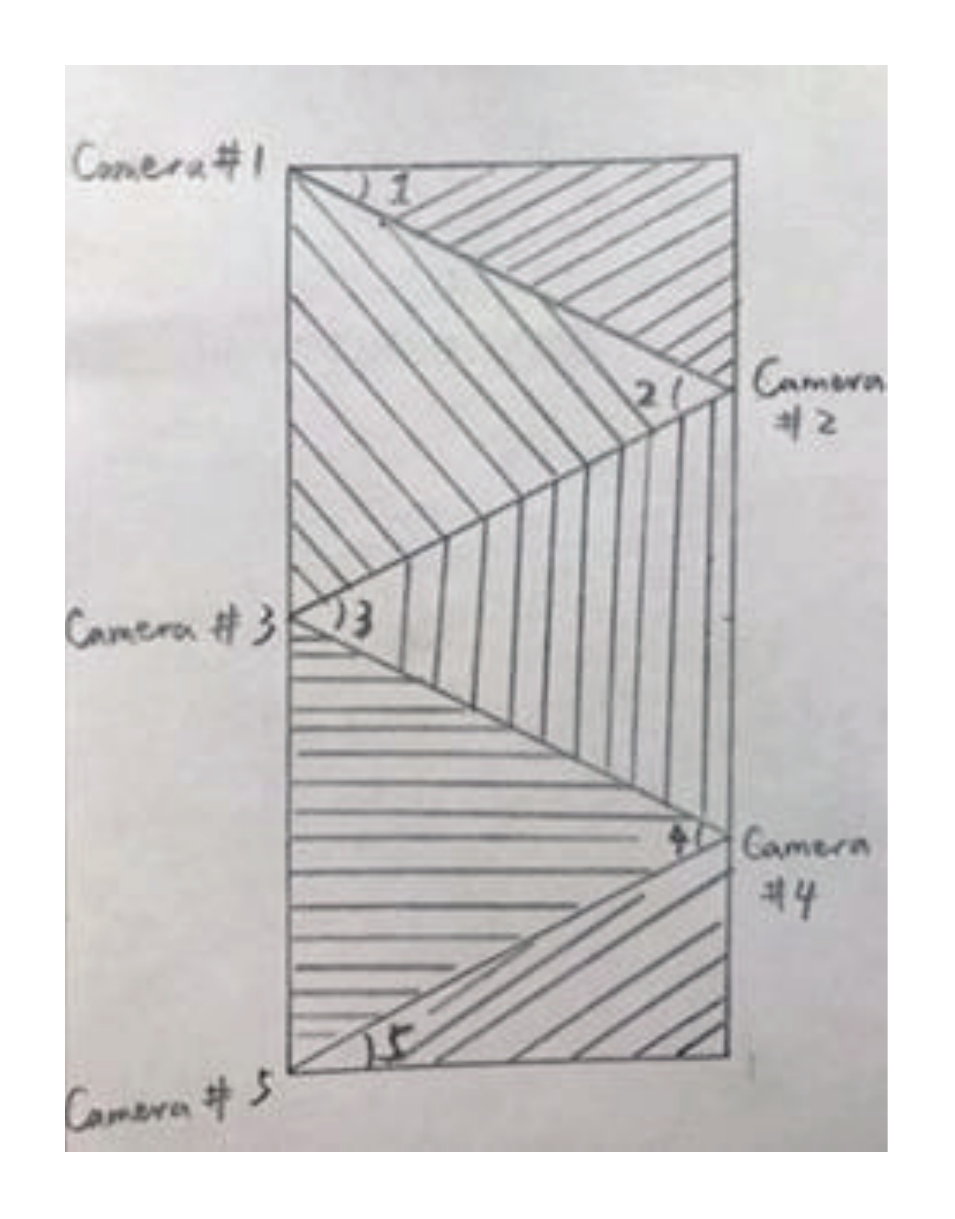}
\end{center}
\caption{Maximum coverage }%
\end{figure}At that time, since all field-of-view angles are in the space, so%
\[
\text{angle}1=\text{angle}2=\text{angle}3=\text{angle}4=\text{angle}%
5=\text{angle}6=2\varphi=2\alpha.
\]
Therefore, AJ // CB // FG // HD// IE. We can see that no fields of view
overlap until the fifth camera is in position, so they maximize the coverage
while no space is counted twice and no space is out of sight. The dashed area
is the fields of view overlap due to the insufficient length of the space, but
it is indeed the best one because applying five cameras will not coverage all
spaces. The dashed area can be more even if we slightly adjust the pose of the
first camera at the corner, and the rest sensors still form sets of a
rectangle. For such an arrangement, the length of a given space, $l$, and the
vertex angle of the sensor, $\alpha$, do matter. The position of the cameras
is determined by them if follows the optimal placements, for which the first
camera should be at one corner of a space, and the others follow, as shown in
Figure 7. Let us name them from top to bottom as number 1, number 2, number 3,
etc.\begin{figure}[h]
\begin{center}
\includegraphics[scale=0.8 ]{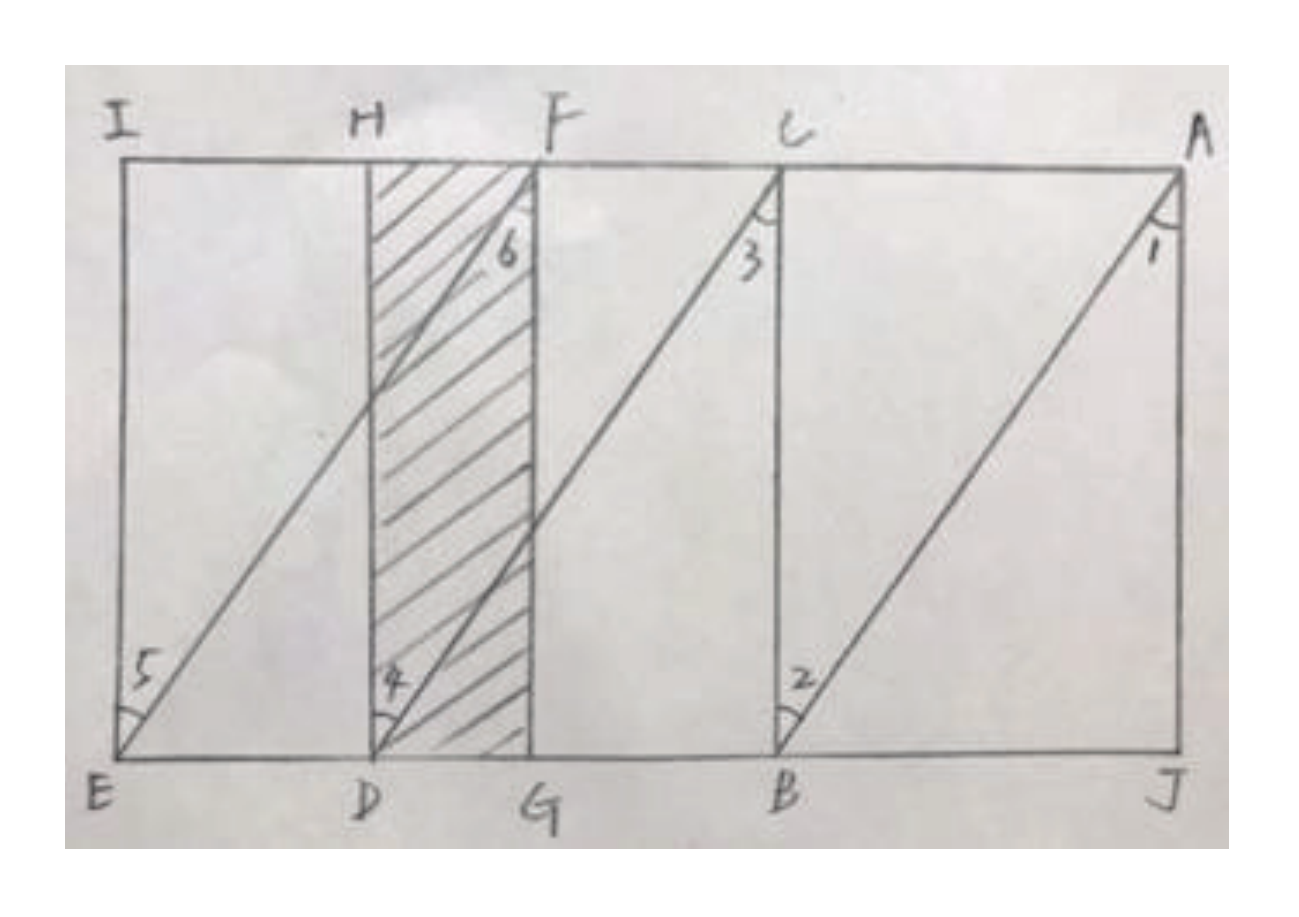}
\end{center}
\caption{Optimal arrangement}%
\end{figure}

Since Camera \#1 is at the corner, where we can see it as the origin (0, 0),
camera \#2's position could be calculated by camera \#1. Let us assume that
$x$ refers to the horizontal position, and $y$ represents the vertical
position of a camera. Then%
\begin{align*}
x  &  =w\\
y  &  =-w\tan \alpha.
\end{align*}
So the position of camera \#2 is $\left(  w,-w\tan \alpha \right)  $, with $y$
position vertically shifts downward at the point where camera \#1's sight edge
intersects that of camera \#2. Similarly, we can calculate other cameras'
positions by each moving downward a distance of $-w\tan \alpha$ and horizontal
positions switch between $0$ and $w$.

\subsection{From the side view}

For analyzing from the side view, it is relatively more straightforward, since
all visual sensors are located on the top of the space. In order to achieve
the maximum coverage, cameras have to be equally spaced so that they can
spread area as much as they can, shown in Figure 8.\begin{figure}[h]
\begin{center}
\includegraphics[scale=0.8 ]{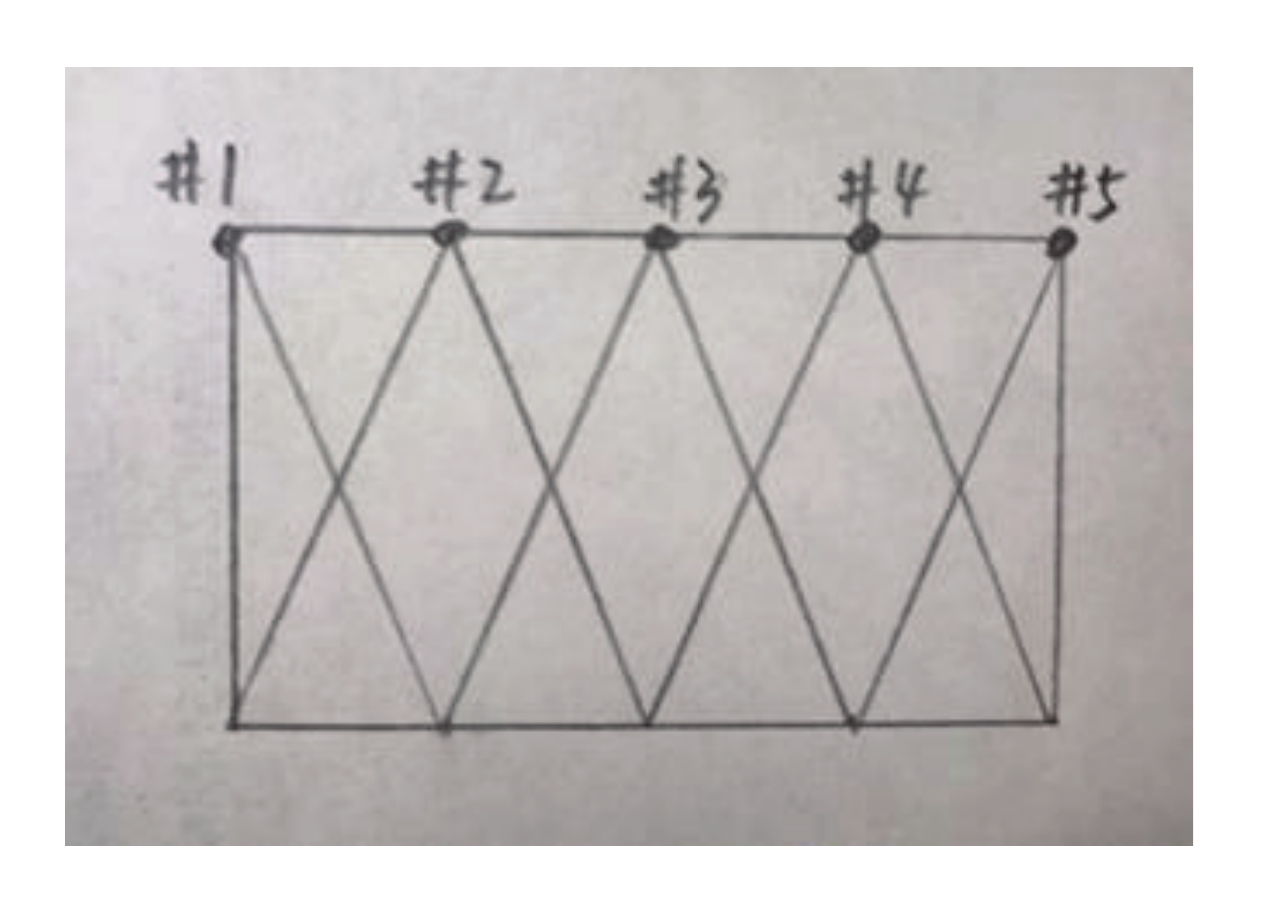}
\end{center}
\caption{Side view}%
\end{figure}

In this case, each camera is separated by a distance of one-fifth of the
length horizontally. Combined the analysis of side view and top view, we can
assume that both are the optimal placements that depend on real situations.

\section{Conclusion and future work}

We have presented an initial study for optimal visual sensor arrangement in a
given space. We have simplified the three-dimension problem to be a
two-dimension problem. However, the camera models are still ideal, and the
space is somewhat simple, such as some variables like the pitch angles of
cameras are not considered. Future works will include these variables and make
it more useful to the practice.


\bigskip

\begin{thebibliography}{9}                                                                                                %


\bibitem {Fradkov(1999)}E. Horster, and R. Lienhart. On the optimal placement
of multiple visual sensors. Proceeding VSSN '06 Proceedings of the 4th ACM
international workshop on Video surveillance and sensor networks, Santa
Barbara, California, USA, October 27, 2006, Pages 111-120.

\bibitem {Shermer(1992)}T.C. Shermer. Recent Results in Art Galleries.
Proceedings of the IEEE, vol. 80, no. 9, 1992, Pages 1384-1399.
\end{thebibliography}
\end{document}